\documentclass[conference]{IEEEtran}
\IEEEoverridecommandlockouts
\usepackage{cite}
\usepackage{amsmath,amssymb,amsfonts}
\usepackage{algorithmic}
\usepackage{graphicx}
\usepackage{textcomp}
\usepackage{hyperref}
\usepackage{xcolor}
\def\BibTeX{{\rm B\kern-.05em{\sc i\kern-.025em b}\kern-.08em
    T\kern-.1667em\lower.7ex\hbox{E}\kern-.125emX}}
\begin{document}

\title{
Comparative Analysis of Transformer Models in Disaster Tweet Classification for Public Safety
% \vspace{-40pt}
}

\author{
\IEEEauthorblockN{\textsuperscript{} Sharif Noor Zisad}
\IEEEauthorblockA{\textit{Department of Computer Science} \\
\textit{University of Alabama at Birmingham}\\
Birmingham, Alabama, USA \\
szisad@uab.edu\vspace{-32pt}}
\and
\IEEEauthorblockN{\textsuperscript{}N. M. Istiak Chowdhury}
\IEEEauthorblockA{\textit{Department of Computer Science} \\
\textit{University of Alabama at Birmingham}\\
Birmingham, Alabama, USA \\
ichowdhu@uab.edu\vspace{-32pt}}
\and
\IEEEauthorblockN{\textsuperscript{}Ragib Hasan}
\IEEEauthorblockA{\textit{Department of Computer Science} \\
\textit{University of Alabama at Birmingham}\\
Birmingham, Alabama, USA \\
ragib@uab.edu\vspace{-32pt}}
}

\maketitle

\begin{abstract}
\label{sec:abstract}
Twitter and other social media platforms have become vital sources of real time information during disasters and public safety emergencies. Automatically classifying disaster related tweets can help emergency services respond faster and more effectively. Traditional Machine Learning (ML) models such as Logistic Regression, Naive Bayes, and Support Vector Machines have been widely used for this task, but they often fail to understand the context or deeper meaning of words, especially when the language is informal, metaphorical, or ambiguous. We posit that, in this context, transformer based models can perform better than traditional ML models. In this paper, we evaluate the effectiveness of transformer based models, including BERT, DistilBERT, RoBERTa, and DeBERTa, for classifying disaster related tweets. These models are compared with traditional ML approaches to highlight the performance gap. Experimental results show that BERT achieved the highest accuracy (91\%), significantly outperforming traditional models like Logistic Regression and Naive Bayes (both at 82\%). The use of contextual embeddings and attention mechanisms allows transformer models to better understand subtle language in tweets, where traditional ML models fall short. This research demonstrates that transformer architectures are far more suitable for public safety applications, offering improved accuracy, deeper language understanding, and better generalization across real world social media text.
\end{abstract}


\section{Introduction}
\label{sec:introduction}

In recent years, social media platforms such as Twitter (currently known as X) have become powerful tools for real time communication during natural disasters, emergencies, and public safety incidents~\cite{seddighi2020saving}. People often use Twitter to share information, report incidents, and express their reactions during critical events like earthquakes, wildfires, floods, or explosions. Automatically detecting such disaster related tweets is important for building systems that support emergency services, early warnings, and public response planning.

Traditional machine learning models have been widely used in text classification tasks, including disaster detection~\cite{kadhim2019survey}. These models, such as Logistic Regression~\cite{lavalley2008logistic}, Naive Bayes~\cite{webb2010naive}, and Support Vector Machines (SVM)~\cite{mammone2009support}, rely on simple representations of text such as Bag-of-Words~\cite{hacohen2020influence} or TF-IDF~\cite{liu2018research}. While these approaches can capture the frequency of words, they ignore the context and deeper meaning behind them. As a result, they often fail to understand informal, figurative, or context dependent language that appears frequently on platforms like Twitter.

\begin{figure}[!ht]
\centering
\includegraphics[width=0.8\columnwidth]{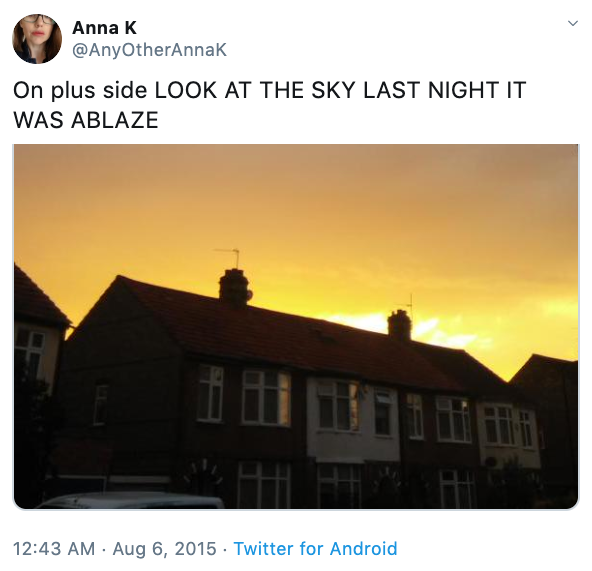}
\vspace{-8px}
\caption{Example of a Tweet which can be misclassified as disaster-related.}
\label{fig:tweet_example}
\end{figure}

For instance, traditional models treat each word independently and are unable to recognize when a word is used metaphorically. If someone writes the word “ABLAZE”, a traditional model may assume that something is literally on fire. However, people often use such language to express excitement or intensity. As an example, we show a sample tweet in Figure \ref{fig:tweet_example}.

\vspace{-10px}
\vspace{-5px}

In this example, a human immediately understands that the word “ablaze” refers to energy or excitement, not a literal fire. This kind of interpretation requires context and world knowledge, which traditional models lack. 

Transformer based models such as  BERT~\cite{devlin2019bert}, DeBERTa~\cite{he2020deberta}, RoBERTa~\cite{liu2019roberta}, and DistilBERT~\cite{sanh2019distilbert} are designed to overcome this limitation. These models use self attention mechanisms to understand the relationships between words in a sentence, allowing them to capture the full context of the message~\cite{rogers2021primer}. As a result, they are capable of identifying the true meaning behind metaphorical or informal expressions ~\cite{zisad2020speech}.

During disasters, traditional news media often struggle to provide detailed and timely updates. In many cases, the infrastructure that supports traditional reporting breaks down or the coverage lacks the fine granularity needed for real time decision making. For instance, during a tornado event that affects an entire state, news agencies could take hours to compile comprehensive updates. In contrast, social media, particularly platforms like Twitter, allows citizens to act as real time reporters, sharing on-the-ground information within minutes. This form of citizen journalism provides a valuable stream of timely, location specific updates. However, not all tweets are relevant, and identifying which ones relate to actual disaster events can be challenging. To accurately filter and classify these posts, we require models that can understand the subtleties and context of human language. Transformer based models are particularly effective in this regard, as they are capable of grasping complex language patterns and delivering higher accuracy in classifying disaster related content.

In this paper, we explore the use of transformer models for disaster tweet classification. We demonstrate that they outperform traditional machine learning approaches by a significant margin. Through a detailed comparison, we highlight the limitations of conventional models and explain why transformer based architectures are better suited for real world public safety applications where understanding sophisticated language terminology is important.
\vspace{5pt} \\
\noindent\textbf{Contribution:}
The contributions of this paper are as follows:
\begin{enumerate}
    \item We systematically evaluated the performance of transformer based models (BERT, DistilBERT, RoBERTa, and DeBERTa) on disaster tweet classification and demonstrated their superiority over traditional machine learning models in terms of accuracy, contextual understanding, and robustness.
    \item We highlighted the key limitations of traditional ML models in handling informal and metaphorical language on Twitter, and showed how transformer models overcome these challenges using self attention and contextual embeddings, making them more suitable for real world public safety applications.
\end{enumerate}

\noindent\textbf{Organization:}
The structure of this paper is as follows: Section \ref{sec:introduction} covers the importance of disaster detection using transformer models, ensuring public safety. Section \ref{sec:background} discusses the background of Tweet Classification, Transformer models and Section \ref{sec:literature_review} reviews the related works. In Section \ref{sec:methodology}, the methodology of our proposed model is outlined along with the methodology. The experiments and the results are reviewed in Section \ref{sec:results}. Finally, Section \ref{sec:conclusion} presents the conclusion and discusses future directions.\\
\vspace{-10pt}
\section{Background}
\label{sec:background}

\subsection{Challenges in Tweet Classification}
Tweets are often short, informal, and unpredictable. People use slang, abbreviations, emojis, and even metaphorical expressions when posting on social media. This makes it difficult to determine whether a tweet is truly about a disaster. For instance, someone might use words like “fire” or “flood” in a figurative sense rather than referring to an actual event. In disaster detection, it is most important that models can tell the difference. Traditional machine learning models fall short in this task since they treat words separately and do not understand their meaning in context. As a result, they may misclassify tweets and generate false alarms, which can weaken the reliability of public safety systems.

Transformer based models help solve this problem. Unlike traditional models, transformers look at the entire sentence and learn how words relate to one another~\cite{rogers2021primer}. They use self-attention mechanisms~\cite{li2020bidirectional} to focus on the most important parts of a sentence, allowing them to understand subtle language cues and variations in meaning. This makes transformers especially effective in processing social media language, where context is the main focus. This precise ability to understand language is very helpful in scenarios where quick and correct decisions are needed, specially when helping people after a disaster.

\subsection{Overview of Transformer Models Used}
In this study, we evaluate four widely used transformer based models: BERT~\cite{devlin2019bert}, DistilBERT~\cite{sanh2019distilbert}, RoBERTa~\cite{liu2019roberta}, and DeBERTa~\cite{he2020deberta}. Each model builds on the original transformer architecture but introduces improvements in training efficiency, performance, or representation quality.

\subsubsection{BERT} Bidirectional Encoder Representations from Transformers, or BERT, was one of the first models to use the power of context in both directions~\cite{devlin2019bert}. During pretraining, BERT randomly “masks” some words in each sentence and learns to predict them (masked language modeling~\cite{salazar2019masked}), while also training to judge whether one sentence logically follows another (next sentence prediction). This dual task forces BERT to build deep, bidirectional awareness of how words relate to one another. In “The roof collapsed after the storm hit”, BERT can understand that the word “storm” is the cause. When finetuned on a specific task, such as classifying disaster tweets, the rich contextual knowledge of BERT allows it to accurately interpret even subtle or metaphorical language.

\subsubsection{DistilBERT} DistilBERT is a lightweight version of BERT, created through knowledge distillation process~\cite{gou2021knowledge}. In distillation, the larger “teacher” model (BERT) teaches a smaller “student” model (DistilBERT) by sharing its internal representations and prediction outputs. The result is a model that retains around 97\% of BERT’s understanding but runs about 60\% faster and occupies 40\% less memory~\cite{sanh2019distilbert}. This reduction in size and increase in speed make DistilBERT an excellent choice for applications that need quick, on-the-fly inference, such as mobile or edge deployments in disaster response systems, without sacrificing much accuracy.

\subsubsection{RoBERTa} Robustly Optimized BERT (RoBERTa) enhances the original BERT framework by eliminating the next sentence prediction objective and concentrating solely on masked language modeling~\cite{liu2019roberta}. Additionally, it employs increased batch sizes, extended sequence lengths, and a more diverse training corpus to achieve superior performance. This streamlined approach allows RoBERTa to learn richer language patterns and yield stronger performance on a wide array of benchmarks. Empirical evaluations frequently demonstrate that RoBERTa surpasses the original BERT in tasks such as sentiment analysis and text classification. Consequently, it is a preferred choice when peak accuracy is required and adequate computational resources are available.

\subsubsection{DeBERTa} Decoding-enhanced BERT with Disentangled Attention, DeBERTa takes the transformer’s self-attention mechanism a step further by separating (disentangling) the representations of content and position ~\cite{he2020deberta}. In other words, DeBERTa learns not only what each word means but also exactly where it sits in the sentence, using relative position encoding~\cite{wu2021rethinking} to capture proximity and order more precisely. This disentangling helps DeBERTa generalize better to new sentence structures and contexts. This is an advantage when dealing with the unpredictable, varied language of social media posts. As a result, DeBERTa often outperforms other BERT variants on tasks that require fine-grained understanding of word relationships.

Each of these models was finetuned on our disaster tweet dataset, allowing us to evaluate their effectiveness in understanding and classifying short, real world text.

\subsection{Importance in Public Safety Context}
In public safety and emergency response scenarios, the ability to rapidly and accurately identify disaster related tweets is essential. Misclassifying a critical tweet as irrelevant, or vice versa, can lead to delays in response or the spread of misinformation. Transformer models offer a more reliable solution because they can distinguish between literal emergencies and metaphorical expressions.

For example, as shown in Figure \ref{fig:tweet_example}, a tweet using the word ``ablaze" in a non-literal context would likely confuse a traditional model. In contrast, a transformer model may derive from surrounding words that the user is describing excitement rather than fire. This level of understanding is important for systems that support first responders, public alert systems, and crisis management dashboards.

By applying transformer models to disaster tweet classification, we move closer to building intelligent systems that can support public safety operations with high accuracy and contextual awareness. The findings in this research highlight how transformer models can be integrated into real time alert pipelines, helping authorities make informed decisions faster and more reliably.

\section{Literature Review}
\label{sec:literature_review}
Recent advances in Natural Language Processing (NLP) have significantly enhanced the ability to analyze social networks for real time disaster monitoring. Traditional machine learning approaches for tweet classification, such as Logistic Regression \cite{lavalley2008logistic} and Naive Bayes \cite{webb2010naive}, typically rely on bag-of-words \cite{hacohen2020influence} or TF-IDF vectorization \cite{imran2015processing}. These methods are computationally efficient but often fall short in capturing context and nuanced meaning, especially within the informal, short text context of Twitter  \cite{kundu2018classification}.

Ray et al. introduced a cross lingual, multilabel classification framework for disaster related tweets using Multilingual BERT (mBERT) \cite{chowdhury2020cross}. The authors compiled a multilingual dataset of over 130,000 samples and employed Manifold Mixup \cite{verma2019manifold} to enhance generalization across languages. Their approach demonstrated improved performance in zero-shot settings, effectively classifying tweets in languages not seen during training. However, the complexity and computational requirements of the model may pose challenges for real-time deployment in resource-constrained environments.

Dinani and Caragea conducted a comparative study that evaluated the performance of Capsule Neural Networks (CapsNets), BERT, LSTM, and Bi-LSTM models on disaster tweet classification tasks~\cite{dinani2023comparison}. Analyzing CrisisBench and CrisisNLP datasets, they found that BERT outperformed other models in both binary and multi class classification tasks. While CapsNets showed competitive results, BERT's superior performance highlighted its effectiveness in capturing contextual information. The study noted that BERT's computational demands could be a limitation for real time applications.

Koshy and Elango proposed a multimodal approach combining text and image data for disaster tweet classification~\cite{koshy2023multimodal}. The system integrated a fine tuned RoBERTa model for text and a Vision Transformer~\cite{han2022survey} for images, employing a multiplicative fusion strategy. After evaluating seven disaster related datasets, the model achieved accuracy rates between 94\% and 98\%. Despite its high performance, this approach depends on the availability of both text and image data. Consequently, its applicability is constrained in scenarios where only a single modality is present.

Basit et al. \cite{basit2023natural} presented a hierarchical classification system for tweets about natural disasters using multimodal data. By aggregating subclasses into several broader categories (Humanitarian, Structure, Non-Informative), the model focused to assist first responders in prioritizing information. The dataset was enriched with tweets from the HumAID dataset to address class imbalance. While the approach improved classification accuracy, the reliance on hierarchical labeling could introduce complexity in real-time scenarios.

Le \cite{le2022disaster} and Chanda \cite{chanda2021efficacy} both found that contextualized embeddings significantly improve disaster tweet classification. Le fine-tuned a BERT model on a Kaggle dataset, achieving 84\% accuracy and demonstrating that bidirectional context captures the subtle language of social media. On the other hand, Chanda compared BERT embeddings with traditional, context free vectors such as GloVe and FastText, finding that classifiers using BERT’s contextual representations consistently outperformed those relying on static embeddings. However, Le recommended incorporating metadata, such as timestamps and user locations to further enhance accuracy, while Chanda noted that BERT’s high computational requirements may limit its use in large scale or real time applications.

\section{Methodology}
\label{sec:methodology}
This section details our experiment, including dataset preparation, system architecture, and evaluation metrics.

\subsection{Dataset}
We have utilized the Kaggle ``NLP Getting Started" competition dataset \cite{kagglenlpgettingstarted}, which contains over 10,000 unique tweets labeled for disaster relevance. Each record includes an identifier, an extracted keyword, an optional user-reported location, the raw tweet text, and a binary target label indicating whether the tweet is related to a disaster. To prepare the data, we first remove any duplicate entries and discard tweets lacking text content. Next, we normalize the remaining tweets by converting all characters to lowercase, stripping out URLs and user mentions, retaining hashtag words while removing the \# symbol, and eliminating punctuation and excessive whitespace. Although the dataset includes metadata fields such as keyword and location, our study focuses exclusively on the tweet text and its associated label. Finally, we split the cleaned dataset into training and testing subsets in an 80:20 ratio using stratified sampling to preserve the original class distribution, resulting in approximately 8,000 training tweets and 2,000 testing tweets.

\subsection{System Architecture}
\begin{figure}[!ht]
    \centering
    \includegraphics[width=1\columnwidth]{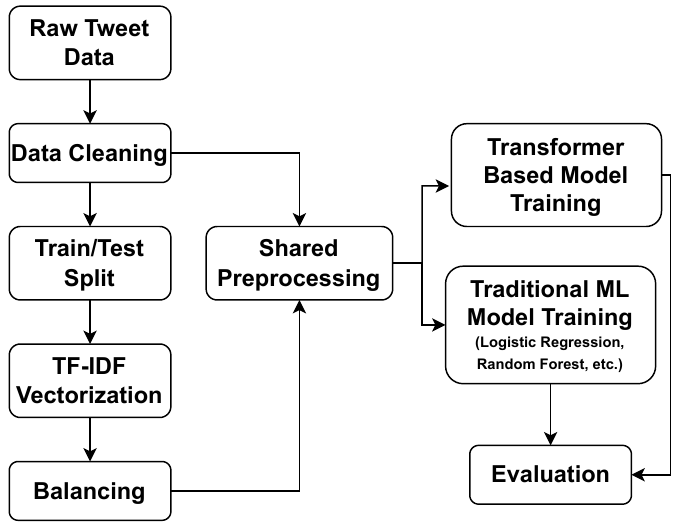}
    \vspace{-20px}
    \caption{System Architecture}
    \label{fig:system_architecture}
\end{figure}

Figure \ref{fig:system_architecture} depicts our end-to-end disaster tweet classification pipeline. 

\vspace{-10px}
\vspace{-5px}

The dataset is a collection of raw tweets derived from the Kaggle dataset, which contains labeled messages collected during various emergency events. Once loaded, these messages pass through a centralized cleaning stage. For instance, all text is converted to lowercase, hyperlinks and user mentions are removed, and extraneous punctuation and whitespace are stripped out. Immediately thereafter, we split the cleaned corpus into training and testing subsets in an 80:20 ratio while preserving the original proportion of disaster versus non-disaster tweets.
After partitioning, a shared preprocessing step prepares the text for both classification approaches. In the traditional pipeline, each tweet is vectorized using the TF–IDF scheme~\cite{liu2018research}, which transforms words into weighted numerical features that emphasize terms unique to disaster-related content. These feature vectors are then fed into classical classifiers, such as Logistic Regression, Support Vector Machines, Random Forests, and other machine learning models for training and evaluation on the test data.
Simultaneously, the transformer pipeline takes the same cleaned tweets and applies Hugging Face tokenization~\cite{jain2022hugging} to produce input IDs and attention masks. We then fine-tune four pretrained transformer models, such as BERT, DistilBERT, RoBERTa, and DeBERTa on the training subset, leveraging their self-attention mechanisms to capture contextual shade that traditional methods miss. After fine-tuning, each transformer outputs predictions on the test set.
Finally, the outputs from both pipelines converge in a unified evaluation stage, where we compare model performance using accuracy, precision, recall, F1-score, and ROC-AUC. By running feature based and transformer based methods side by side, we enable a direct comparison that clearly demonstrates how modern transformer architectures through their contextual understanding outperform traditional text classification techniques.

\subsection{Evaluation Metrics}
To evaluate each model’s effectiveness, we applied a range of established classification metrics to the test set. We began with accuracy, which quantifies the proportion of tweets that the model labels correctly. The equation to calculate accuracy is as follows:

\begin{equation}
\label{eqn:accuracy}
\text{Accuracy} = \frac{TP + TN}{TP + TN + FP + FN}
\end{equation}

Where $TP$, $TN$, $FP$, and $FN$ denote true positives, true negatives, false positives, and false negatives, respectively. Precision then measures the fraction of tweets flagged as disasters that are indeed disaster related, while recall assesses the proportion of actual disaster tweets that the model successfully identifies. Equation ~\ref{eqn:precision} and ~\ref{eqn:recall} are used to calculate precision and recall.

\begin{equation}
\label{eqn:precision}
\text{Precision} = \frac{TP}{TP + FP}
\end{equation}

\begin{equation}
\label{eqn:recall}
\text{Recall} = \frac{TP}{TP + FN}
\end{equation}

\noindent
Since precision and recall often trade off against one another, we also calculated the F1-score, the harmonic mean of precision and recall, which gives a single, balanced indicator of performance under imbalanced class distributions. The algorithm to calculate F1-score is presented in equation \ref{eqn:f1-score}:

\begin{equation}
\label{eqn:f1-score}
F1-Score = 2 \times \frac{\text{Precision} \times \text{Recall}}{\text{Precision} + \text{Recall}}
\end{equation}

\noindent
To understand how well each model distinguishes between disaster and non-disaster tweets across all possible decision thresholds, we compute the area under the receiver operating characteristic curve (ROC-AUC). The ROC curve plots the true positive rate (TPR) against the false positive rate (FPR) at various decision thresholds, where

\begin{equation}
\text{TPR} = \frac{TP}{TP + FN}, 
\quad
\text{FPR} = \frac{FP}{FP + TN}
\end{equation}

\noindent
The ROC–AUC score summarizes this curve as a single value in $[0,1]$, with higher values indicating better discriminative ability across thresholds.
\section{Results and Discussion}
\label{sec:results}

We have trained the dataset using six traditional machine learning models and four transformer models. The machine learning models are Logistic Regression (LR), Support Vector Machines (SVM), Random Forests (RF), Gradient Boosting (GB), Naive Bayes (NB) and Extreme Gradient Boosting (XGBoost). Table \ref{tab:results_ml} reports the performance of classical machine-learning models.
\vspace{-10px}
\begin{table}[!ht]
    \small
    \centering
    \caption{Experimental Results for Traditional ML Models}
    \label{tab:results_ml}
    \begin{tabular}{|c|c|c|c|c|}
    \hline
     \textbf{Model} & \textbf{Accuracy} & \textbf{Precision} & \textbf{Recall}  & \textbf{F1-Score} \tabularnewline
    \hline
      LR & 0.82 & 0.82 & 0.81 & 0.81 \\
    \hline
      SVM & 0.80 & 0.80 & 0.80 & 0.80 \\
    \hline
      RF & 0.79 & 0.80 & 0.77 & 0.78 \\
    \hline
      GB & 0.76 & 0.77 & 0.74 & 0.74 \\
    \hline
      NB & 0.82 & 0.83 & 0.80 & 0.81 \\
    \hline
      XGBoost & 0.79 & 0.79 & 0.78 & 0.78 \\
    \hline
    \end{tabular}
\end{table}

According to the result in Table \ref{tab:results_ml}, Logistic regression (LR) and Naive Bayes (NB) achieved the highest overall accuracy at 82 percent. LR achieved a precision of 0.82 and recall of 0.81, while NB had slightly higher precision at 0.83 but similar F1-score of 0.81 as the recall is lower (0.80). Support vector machines (SVM) followed closely with balanced precision, recall, and F1-score of 0.80 across the board. Random forests (RF) attained 79 percent accuracy, demonstrating solid precision (0.80) but somewhat lower recall (0.77), resulting in an F1-score of 0.78. XGBoost also achieved 79 percent accuracy, with precision and recall of 0.79 and 0.78 respectively. Gradient boosting (GB) lagged behind the other models, with 76 percent accuracy and the lowest recall at 0.74, having an F1-score of 0.74. Overall, while traditional methods can reach respectable performance, their metrics cluster around the 80s at best, underscoring the need for more context-aware approaches. 
The experimental result of transformer models is depicted in Table \ref{tab:results_transformer}.

\vspace{-10px}
\begin{table}[!ht]
    \small
    \centering
    \caption{Experimental Results for Transformer Models}
    \label{tab:results_transformer}
    \begin{tabular}{|c|c|c|c|c|}
    \hline
     \textbf{Model} & \textbf{Accuracy} & \textbf{Precision} & \textbf{Recall}  & \textbf{F1-Score} \tabularnewline
    \hline
      DistilBERT & 0.90 & 0.90 & 0.90 & 0.90 \\
    \hline
      RoBERTa & 0.84 & 0.84 & 0.84 & 0.84 \\
    \hline
      DeBERTa & 0.83 & 0.83 & 0.83 & 0.83 \\
    \hline
      BERT & 0.91 & 0.91 & 0.91 & 0.91 \\
    \hline
    \end{tabular}
\end{table}

Based on the experimental results presented in Table \ref{tab:results_transformer}, BERT achieved the highest performance across all metrics with an accuracy, precision, recall, and F1-score of 0.91, indicating it is the most effective at correctly identifying disaster-related content. DistilBERT closely followed with 0.90 across all metrics, showing that it can deliver near-BERT performance with significantly reduced computational cost, making it suitable for real time or edge deployment. RoBERTa achieved moderate performance with 0.84 in all metrics, while DeBERTa, often known for strong benchmarks, achieved a slightly lower score of 0.83 in this setup. This suggests that while all models are effective, BERT remains the most accurate. 

\begin{figure}[!ht]
    \centering
    \includegraphics[width=1\columnwidth]{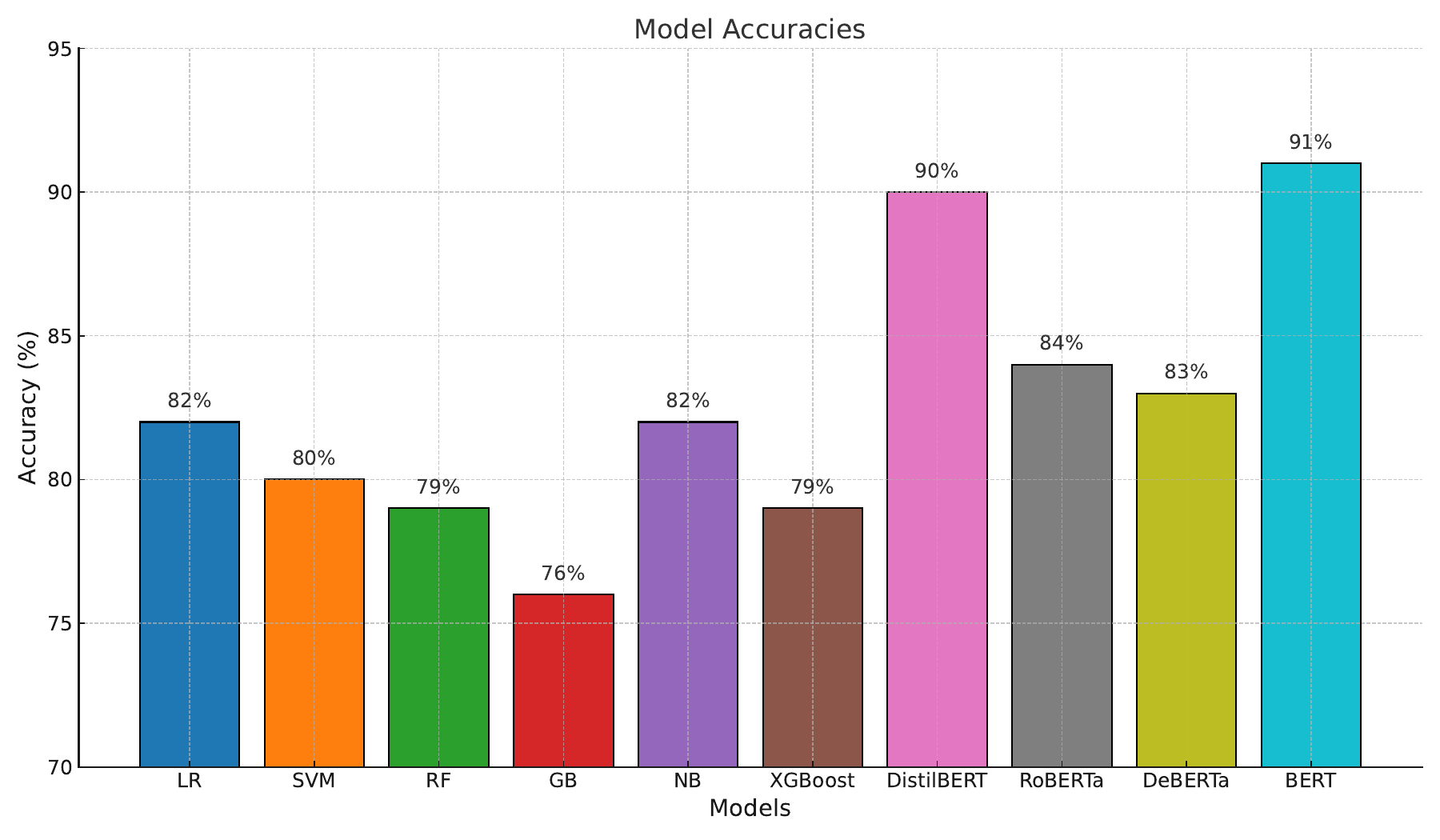}
    \vspace{-20px}
    \caption{Performance Comparison}
    \label{fig:performance_comparison}
\end{figure}

Figure \ref{fig:performance_comparison} shows the performance (accuracy) comparison of both traditional machine learning models and transformer models. 

\vspace{-10px}
\vspace{-5px}

According to the performance comparison plot in Figure \ref{fig:performance_comparison}, the accuracy of traditional methods, such as Logistic Regression, SVM, Random Forest, Gradient Boosting, Naive Bayes, and XGBoost, is in between 76\% and 82\%, with Logistic Regression and Naive Bayes peaking at 82\% and Gradient Boosting at the lower end with 76\%. On the other hand, transformer models deliver substantially higher accuracy. DistilBERT reaches 90\%, RoBERTa 84\%, DeBERTa 83\%, and BERT tops the list at 91\%. This clear gap highlights how transformer architectures, which capture context and word relationships, outperform feature based approaches in classifying real world social media text. While BERT outperforms DistilBERT by only one percentage point, as we discussed in Section \ref{sec:background}, DistilBERT’s faster inference speed and reduced resource requirements make it an ideal choice for public safety deployments~\cite{zisad2024towards}. For this reason, we selected DistilBERT for our subsequent analyses.

\begin{figure}[!ht]
    \centering
    \includegraphics[width=1\columnwidth]{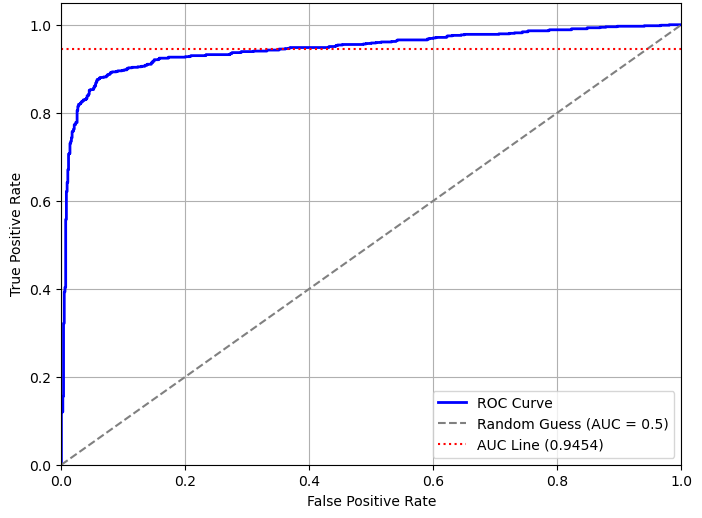}
    \vspace{-20px}
    \caption{ROC Curve}
    \label{fig:roc_curve}
\end{figure}

The ROC curve for the DistilBERT classifier is presented in Figure \ref{fig:roc_curve}.

\vspace{-10px}
\vspace{-5px}

The ROC curve shows that the model performs very well in distinguishing between disaster related and non-disaster tweets. The curve rises sharply toward the top left corner, which means that the model has a high true positive rate and a low false positive rate. The AUC score of 0.9454 indicates excellent overall performance, meaning the model correctly ranks most disaster tweets higher than non-disaster ones. This suggests that the model is reliable and effective for use in public safety systems where accurately identifying emergency information from social media is important.
\section{Conclusion}
\label{sec:conclusion}
In this work, we have demonstrated that transformer based models substantially outperform traditional machine learning methods in the task of disaster tweet classification. Our experiments show that, while classical algorithms such as Logistic Regression and Naive Bayes reach up to 82\% accuracy, transformer models, including DistilBERT, RoBERTa, DeBERTa, and BERT achieve higher performance, with BERT leading at 91\% and DistilBERT closely following at 90\%. DistilBERT, in particular, keeps a balance between computational efficiency and classification accuracy, making it an ideal choice for real time public safety applications.

The parallel pipeline design employed in this study enables a rigorous, side-by-side comparison of traditional and transformer based methods, while highlighting the critical role of contextual embeddings and self-attention in processing informal, nuanced social media text. By modeling relationships between words and incorporating broader semantic context, transformer architectures significantly reduce the misclassification errors that often occurs in bag-of-words or TF-IDF systems. These results indicate that emergency management agencies and public safety organizations would benefit from deploying transformer based models in automated disaster monitoring platforms to achieve faster, more accurate situational awareness and improve response times.

In our future research, we can explore the incorporation of auxiliary metadata, such as user location, temporal features, and image content to further enhance model robustness. Moreover, expanding this framework to multilingual tweet streams and low resource languages would further broaden its applicability in global crisis scenarios. Overall, this study provides a clear roadmap for leveraging transformer models in public safety analytics and make the foundation for more advanced, context aware emergency detection systems.

\bibliographystyle{IEEEtran}
\bibliography{references}

\end{document}